\documentclass{article}

\usepackage{PRIMEarxiv}

\usepackage[utf8]{inputenc} 
\usepackage[T1]{fontenc}    
\usepackage{hyperref}       
\usepackage{url}            
\usepackage{booktabs}       
\usepackage{amsfonts}       
\usepackage{nicefrac}       
\usepackage{microtype}      
\usepackage{lipsum}
\usepackage{fancyhdr}       
\usepackage{graphicx}       
\graphicspath{{media/}}     
\usepackage{amsmath} 
\usepackage{algorithm}
\usepackage{algpseudocode}
\usepackage{multirow}
\usepackage{subfig}

\usepackage{footnote}
\usepackage{authblk}

\newcommand*{\affaddr}[1]{#1}
\newcommand*{\affmark}[1][*]{\textsuperscript{#1}}

\usepackage[english]{babel}
\usepackage[doi=false,isbn=false,url=false,eprint=false]{biblatex}

\AtEveryBibitem{
    \clearfield{urlyear}
    \clearfield{urlmonth}
}

\title{MultiSlot ReRanker: A Generic Model-based Re-Ranking Framework in Recommendation Systems
}

\author{
    Qiang Charles Xiao \affmark[1, $\dagger$]  
    Ajith Muralidharan \affmark[1, $\dagger$]  
    Birjodh Tiwana  \affmark[$\dagger$]   
    Johnson Jia  \affmark[$\ddagger$] \thanks{Work performed while at LinkedIn.} 
    \ Fedor Borisyuk  \affmark[$\dagger$]   
    Aman Gupta  \affmark[$\dagger$]  
    Dawn Woodard  \affmark[$\dagger$]  \\
    \affaddr{\affmark[$\dagger$] LinkedIn} \\ \affaddr{\affmark[$\ddagger$] Google}  
}

\date{}

\begin{document}
\maketitle

\footnotetext[1]{Primary authors.}

\begin{abstract}
In this paper, we propose a generic model-based re-ranking framework, MultiSlot ReRanker, which simultaneously optimizes relevance, diversity, and freshness. Specifically, our Sequential Greedy Algorithm (SGA) is efficient enough  (linear time complexity) for large-scale production recommendation engines. It achieved a lift of $+6\%$ to $ +10\%$ offline Area Under the receiver operating characteristic Curve (AUC) which is mainly due to explicitly modeling mutual influences among items of a list, and leveraging the second pass ranking scores of multiple objectives. In addition, we have generalized the offline replay theory to multi-slot re-ranking scenarios, with trade-offs among multiple objectives. The offline replay results can be further improved by Pareto Optimality. Moreover, we've built a multi-slot re-ranking simulator based on OpenAI Gym integrated with the Ray framework. It can be easily configured for different assumptions to quickly benchmark both reinforcement learning and supervised learning algorithms. 
\end{abstract}


\section{Introduction}
In general, recommendation systems \cite{belloSeq2SlateRerankingSlate2019a} \cite{covingtonDeepNeuralNetworks2016} consist of three stages: 1) retrieval; 2) second pass ranking; 3) re-ranking. The re-ranking stage is critical, as it directly affects users' experience, while satisfying business constraints. Typical re-ranking strategies include: 1) filtering items by rules; 2) modifying the second pass ranking score by some criteria such as item age (freshness); 3) slotting rules such as placing promotion items to fixed slot positions; 4) rules such as minimum distance of similar items in terms of creators, type and content (diversity). 

Some re-ranking rules are necessary due to business requirements. However, the system will soon become too complex and hard to maintain, with more and more rules added in re-ranking stage. Moreover, hard-coded rules may not be optimal, limiting the potential of personalization in recommendation systems. Recently, the industry has been paying more attention to multislot or whole page re-ranking.

Although pairwise and listwise learning to rank approaches consider a pair or list of items as input, they only optimize for loss functions without explicitly modeling the
mutual influences among items in the feature space. In addition, they do not capture long term impact of whole page re-ranking across multiple user sessions. 

In this paper, we propose a generic model-based re-ranking framework, MultiSlot ReRanker, to optimize relevance, diversity and freshness simultaneously. The framework explicitly models the mutual influences among items, and leverages the second pass ranking scores of multiple objectives. 

\textbf{Our main contributions} are:
\begin{itemize}
    \item A multi-slot re-ranking algorithm, the Sequential Greedy Algorithm (SGA), is efficient enough for large-scale production recommendation engines (linear time complexity). 
    \item We have generalized the offline replay theory to multi-slot re-ranking scenarios. The offline replay results can be further improved by Pareto Optimality.
    \item We have built a recommendation system simulator based on OpenAI Gym and scaled it with the Ray framework. The simulator can quickly benchmark both reinforcement learning and supervised learning algorithms.
\end{itemize}

\section{Related Work}
There are mainly three approaches to solve the multislot recommendation problems: 1) Modeling approaches that sequentially construct the slate functions \cite{belloSeq2SlateRerankingSlate2019a} \cite{dingWholePageOptimization2019a}; 2) Modeling approaches that are based on whole page scoring functions \cite{jiangGreedyRankingSlate2019} \cite{peiPersonalizedRerankingRecommendation2019a}; 3) Modeling approaches that also model long term impact of slate construction \cite{ieSlateQTractableDecomposition2019a} \cite{zhaoDeepReinforcementLearning2018a}. However, these approaches either have high latency thus become impractical in large scale industrial recommendation systems, or lack the capability of specifically modeling the mutual influences among items in a list.

List-CVAE \cite{jiangGreedyRankingSlate2019} is a generative model for the entire slate construction. It models the distribution of items in the state that corresponds to a given reward. During encoding, we have a context (expected response) and the ordered state. This is compressed to an embedding, which is conditioned on the expected response. It also models the context prior. During scoring, given the expected response, the decoder decodes the state. Since we don't know how the user will respond, the expected response during decoding is setup as $(1,1,...)$ (i.e., click of every item). This may not be achievable. During decoding, it computes dot product of document embeddings with the final layer to decode the different items in different positions in the slate. It may decode sequentially to prevent items from being duplicated.

A page wise re-ranking layer is proposed in \cite{peiPersonalizedRerankingRecommendation2019a} to account for interactions among items and also user interactions. The output is still a softmax layer predicting whether the user will click on an item or not in this session. The entire list, ordered by scores from the previous ranking layer, is fed into an attention network after featurizing with embeddings including personalized embeddings and position-aware components. The network outputs the $p(utility)$ score for each item. The items are re-ranked according to the output scores. 
However, the context which serves as the input to the model will not be consistent with the context after re-ranking. Thus, the re-ranking step may not be guaranteed to work. It is evaluated both offline and online successfully to show gains against pointwise ranking. However, this is no comparisons against sequence approaches such as Seq2Slate \cite{belloSeq2SlateRerankingSlate2019a} because of latency constraints.

A constrained optimization problem is formulated in \cite{dingWholePageOptimization2019a} to maximize primary objectives (e.g., Clicks) while satisfying diversity constraints (e.g., total number of impressions in certain category). Impact of diversity is modeled in both primary objectives and the constraints. It is applicable to Amazon video where there are ranking carousels with different category (e.g., subscription video, transactional video, third party channels). It assumes that a user scans videos from top to bottom, and slot is decided based on updates in slot above. The greedy sequential optimization procedure uses primal dual algorithms to satisfy constraints. It has been shown that diversity features have impact (AUC increases by $4\%$) when using the models without constraints. Models with constraints are also working as intended.

A sequential scoring framework is proposed in \cite{belloSeq2SlateRerankingSlate2019a} where each item's score depends on the previous items chosen. It is a sequence to sequence approach. First, all items are sequentially consumed to get a comprehensive understanding of the set of items in a state, i.e., its encoding. Then, items are sequentially decoded, where decoding depends on the previous state (i.e., the item chosen during a decoding step serves as an input to the next step). However, the computational complexity is $O(N^2)$. It uses pointer networks which points to the item during the encoding process. It borrows reinforcement learning methodology (e.g., REINFORCE) to optimize the problem since good labels are not directly seen. However, REINFORCE being a policy gradient method has high sample requirements. It also introduces approximation to simulate a better training process. The loss function is intuition driven (synthetically restricting samples). In benchmark offline simulation data, it shows $15\%$ increase in precision against strong baselines. Almost all algorithms are better than baseline. The one-step decoder seems to work on simulation data. Results on real data indicates mean average precision (MAP) increase of $20-30\%$ and click-through rate (CTR) absolute increase of 0.5 to 1. The one-step decoder seems to also work in practice, with some suboptimal results. It mostly uses the greedy decoder to manage sampling complexity.

The DeepPage \cite{zhaoDeepReinforcementLearning2018a} optimizes a page of items with proper display based on real-time feedback from users. It aims to generate a set of complimentary items, and ranks items into a 2D grid page instead of 1D list. It models the sequential interaction between users and the platform as a Markov decision process (MDP) process. More specifically, the state is defined as users' preferences, and action is defined as one page of items. There are two challenges: 1) action space is dynamic and huge; 2) item indices cannot model relation between different items. The solution is based on the Actor-Critic framework. The encoder generates current state based on item embedding representation, item category, and users’ feedback.
The action generation uses Deconvolution Neural Network to restore the page of item embeddings from the encoded state, then maps to low-dimension dense vector through 2D-CNN.

The SlateQ \cite{ieSlateQTractableDecomposition2019a} addresses challenges of recommending multiple items simultaneously where the presence of one item can influence the user response on other items. It develops a slate decomposition technique that estimates the long term value of a whole slate of items by estimating the long term value of the individual items that comprise the slate, i.e., decompose slate Q-values into item Q-values (conditioned on user click). The value of the slate depends on a user choice model. The assumptions are as follows: 1) User consumes a single item from each slate, which may be null item; 2) Reward depends only on the item consumed by the user (not other items in slate). The state transition also depends only on the consumed item. The Q value of slate is defined as reward associated with reward at the state with the slate + discounted future reward. With two assumptions above, the Q value decomposes to $P(click) * item-wise Q value$, where $P(click)$ is user behavior model. The general user choice model does not consider position in calculating the probability a user selects an item in a slate. It demonstrates the details of how this optimization problem can be expressed as a fractional linear program, which can then be optimized. At serving time, an approximation for constructing the slate is to pick the top $k$ items with highest score in descending order. 

\section{MultiSlot ReRanker Framework}
Given a ranked list of items along with the pointwise utility model scores, we aim to set up a re-ranking layer that produces an ordered list of items. The objective is to optimize joint metrics across the entire list while utilizing the pointwise scores as a crucial component of the re-ranking layer. 

The MultiSlot ReRanker framework explores strategies to adhere to practical scoring latency considerations. The re-ranking scoring complexity can be tuned to be of the order of $O(K*N)$ (with $K < N$ being a tuning parameter, $N$ is the total number of items scored in the list) so that it is more efficient than the re-ranking algorithms seen in literature which require $O(N^2)$ scoring operations.

\begin{figure}
  \centering
  \includegraphics[scale=0.5]{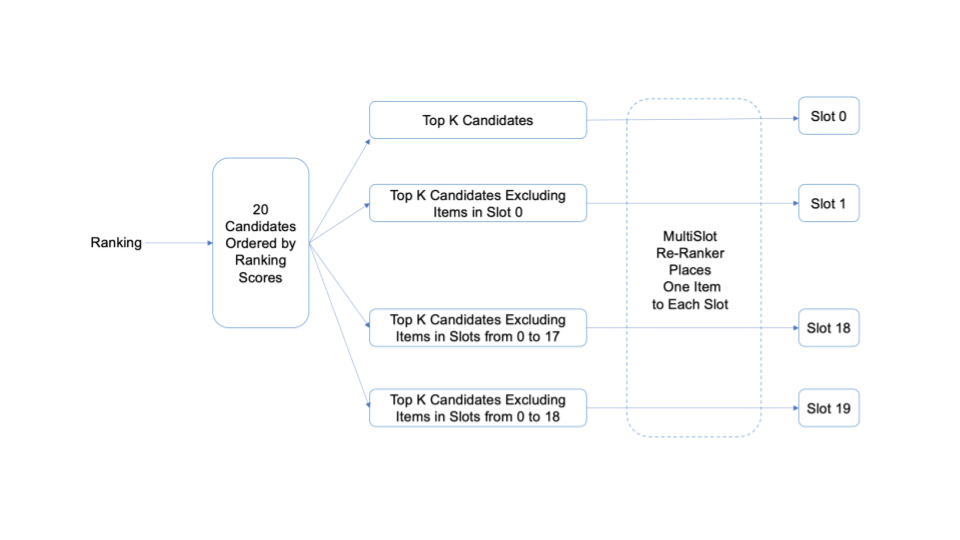}
  \caption{The MultiSlot ReRanker Framework (e.g., with 20 Slots).}
  \label{fig:multislot}
\end{figure}

This re-ranking layer is called MultiSlot ReRanker framework as shown in Figure \ref{fig:multislot}. The list construction will proceed sequentially from the first item in the list. Scoring at each step (or slot) can utilize information about items which have been slotted above as well as candidate items remaining in the re-ranking process. For example, assuming there are 20 slots in total, the MultiSlot ReRanker framework will output 20 items. The input is a list of 20 items ordered by second pass ranking (SPR) scores. At slot 0, it selects top $K$ candidates from 20 items, then place one candidate to slot 0. At slot 1, it selects top $K$ candidates from remaining 19 items, then place one candidate to slot 1. This process is continued until all items are placed on all slots.    

In addition to the a re-ranked list, we also produce improved estimates of the item level rewards/responses which could be used for further optimization and re-ranking downstream as required. 

\subsection{Sequential Greedy Algorithm}
Going forward, we will use the notation $i$ to represent the position (i.e., slot index). The MultiSlot ReRanker is given an ordered list of $N$ items based on the scores from the ranking layer in previous stage. At each slot, we will allow the MultiSlot ReRanker to take action $a \in\{0 \ldots K-1\}$ where $K$ is the maximum number of candidates to evaluate at any step during re-ranking. Taking an action equal to $k$ would mean that we choose the $k$-th item in the remaining candidates, which are always sorted by scores from the previous layer. This allows us to parameterize the action space. 

The sequential greedy algorithm \ref{alg:sequential_greedy} uses a response prediction model $f$ which predicts the user's response in a slot conditioned on the items in the previous slots. We use that to rank different actions ($a \in\{0 \ldots K-1\}$) and choose the best action $p$.

\begin{algorithm}
\caption{Sequential Greedy Algorithm}\label{alg:sequential_greedy}
\begin{algorithmic}
\Require $i \geq 0$ and $i < N$ \Comment{i is slot or position index, N is total number items.}
\State \texttt{Select item with top SPR score and place it to slot 0}
\State $i \gets 1$ \Comment{The slot 0 does not have previous slot}
\State $C \gets \{0 \ldots K-1\}$  \Comment{Initialize top K candidate set by second pass ranking scores}
\While{$i < N$}
      \State $reranking\_score\_max \gets -\infty$
      \State $k \gets 0$ \Comment{Select item k from Candidates}
      \For{$a \in C$}
        \State \texttt{Extract features from item $a$}
        \State \texttt{Extract interaction features between item $a$ and the previous slot $i-1$}
        \State \texttt{Calculate score by equation \ref{eqn:model_f}}
        \If{$score$ > $reranking\_score\_max$}
            \State $reranking\_score\_max \gets score$
            \State $k \gets a$ \Comment{Pick the item with largest re-ranking score, i.e., greedy}
        \EndIf
      \EndFor

      \State \texttt{Remove item $k$ from Candidate set $C$ and place it to slot $i$}
      \State \texttt{Pick the next top item in remaining list and add it to top K Candidate set $C$}
      \State $i \gets i + 1$
\EndWhile
\end{algorithmic}
\end{algorithm}

As shown in data analysis as well as MultiSlot Simulator performance comparison, "interaction features" among the current slot and previous slots play an important role to achieve better re-ranking and user experience. Thus, the generic model function $f$ for the current slot $i$ is defined as:
\begin{equation}\label{eqn:model_f}
f(second\ pass\ ranking\ score,\ current\ slot's\ features,\ interaction\ features)
\end{equation}
The model function $f$ can be logistic regression, tree\cite{chenXGBoostScalableTree2016}, or deep neural network models depending on the use cases.

\subsection{Offline Replay}
Given the data from a random bucket, which randomizes the order of updates shown to the user, we will use it to replay the effect of the modified ranker changes in the ecosystem. For single slot offline replay, when the model prediction match the random bucket data log\cite{liUnbiasedOfflineEvaluation2011}, it will be considered as positive (click or contributions).   

For multislot/whole page offline replay, importance sampling provides us with the mechanism to achieve this. Assume that the replay data is represented as follows:
\begin{itemize}
\item Slot is represented by subscript $i$
\item Context (or session, which groups the ordered set of items) is represented by subscript $m$ (total number of sessions). Each context is a session where the recommendations are shown.
\item Action corresponding to the slot $i$ is represented by $a_{m, i}^r$, while the action taken by the policy/model given the same set of items above is represented by $a_{m, i}^p$. The probability distributions of those actions are represented by $\pi^r(a)$ and $\pi^p(a)$ respectively. For random bucket $\pi^r(a)$ is uniform. 
\item Reward/user response, as obtained from the random bucket is represented as $r_{m, i}$
\end{itemize}

Note, the reward may be multidimensional, and this definition will extend easily there. The most generic unbiased estimator for the reward from a generic policy is given by 
\begin{equation}
\sum_m \sum_i r_{m, i} w_{m, i}
\end{equation}
, where $w_{m, i}$ is the weight assigned to the collected reward at slot $i$. An unbiased but high variance estimator\cite{chenTopKOffPolicyCorrection2019} for the reward can be obtained by choosing 

\begin{equation}
w_{m, i}=\Pi_{j=1 . i} \pi^p\left(a_{m, j}^r\right) / \pi^r\left(a_{m, j}^r\right).
\end{equation}

For deterministic policies, the numerator is one when all actions made by the random bucket match the actions made by the policy/model. The variance of this estimator increases exponentially with the number of slots, so its use is impractical, unless you restrict the slots to very few, or collect copious amounts of data.

A slightly biased, but low variance estimator is the one-step importance sampling method. This is obtained by choosing
\begin{equation}
w_{m, i}=\pi^p\left(a_{m, i}^r\right) / \pi^r\left(a_{m, i}^r\right)
\end{equation}
, which matches actions just for one step (independently). This has been a very useful estimator in other multi-step offline evaluation scenarios.

The one-step importance sampling offline replay is validated in both MultiSlot Simulator and production dataset. For example, always choosing the second best item out of $K$ candidates will have smaller reward compared to always choosing the best item. 

\section{LinkedIn Feed}
With LinkedIn Feed as an example use case, a partial list of possible features as input of the model function $f$ is shown in Table \ref{table:features}. There are trade-off between two categories of responses: 1) click, i.e., whether the user clicks a feed item; 2) contributions, i.e., whether the user like/comment/share/skip on a feed item. The "skip" can be defined as an user spends less than several seconds viewing a feed item. The "previous slots" could denote all or a portion of previous slots from slot $0$ to slot $i-1$, depending on the trade-off between latency and performance. 

For each response such as click/like/comment/share/vote, an example of re-ranking model $f$ as a logistic regression model can be defined as 
\begin{equation} \label{eqn:model_lr}
{logistic}\left(w ^ { T } * \left[
{ logit }({score}), {itemtype}_{i}, {itemtype }_{i-1}, {itemtype }_{i} * {itemtype }_{i-1}, \\
typecount_{1}, \ldots, typecount_{T}
\right]\right),
\end{equation}
where input features are
\begin{itemize}
    \item $score$ is the utility score output of the second pass ranking model. 
    \item $w$ is the re-ranker model parameter to be estimated during model training.  
    \item ${itemtype}_{i}$ is the type of a candidate LinkedIn Feed item to be placed at the current slot $i$.
    \item ${typecount}_{i}$ is the total count of LinkedIn Feed items type (such as video, ads, job, company, article, etc.) among previous slots from $1$ to $i - 1$
    \item $T$ is the total number of Feed items types. 
\end{itemize}

Note that the spr score feature needs to be converted by $logit$. The $logit$ and $logistic$ functions are defined as
\begin{equation}
{logit}(p)=\log \left(\frac{p}{1-p}\right)
\end{equation}

\begin{equation}
logistic(x)=\frac{1}{1+e^{-x}}.
\end{equation}

Once the re-ranking model $f$ is individually trained on click/like/comment/share/skip responses, the predicted probabilities of each responses are combined with a formula to generate the final re-ranking score used in Algorithm \ref{alg:sequential_greedy}.

\begin{table}
\caption{A List of Features for Sequential Greedy Algorithm}
    \begin{center}
    \begin{tabular}{ |c|l| } 
    \hline
    Feature Category & Features \\
    \hline
    \multirow{3}{6em}{Second Pass Ranking Scores} & p(Click)  \\ 
    & Contribution responses such as p(Like), p(Comment), p(Share), p(Skip) \\ 
    & p(Contributions) whose label is positive if any of Like, Comment, Share, Skip is positive. \\ 
    \hline
    \multirow{3}{6em}{Current Slot’s Features} & Slot index $i$ \\ 
    & Embeddings of item at slot $i$ \\ 
    & Type of item at slot $i$ such as Video, Image, Job, Company, Article, etc. \\ 
    \hline
    \multirow{3}{6em}{Interaction Features} & Type of items in previous slots \\ 
    & Count of each type of items in previous slots \\ 
    & Cross feature with item type among slot $i$ and previous slots \\ 
    & Embeddings dot product of items among current slot $i$ and previous slots \\ 
    & Whether items at current slot $i$ and previous slots are created by the same user \\ 
    \hline
    \end{tabular}
    \end{center}
\label{table:features}
\end{table}

\subsection{Data Analysis}
We've done both offline and online opportunity analysis for MultiSlot ReRanker in LinkedIn Feed. For example, we calculated the ratio of actual click through rate to predicted click through rate at slot 2, based on the item type at slot 0. The item type includes IMAGE, VIDEO, Activity, Company, and Job. The actual ratio is omitted here because of business confidentiality.


We've also done online A/B test to measure the mutual interaction among feed items. For example, there are several re-rankers in feed models that are meant to ensure a diverse set of items is shown in each feed sessions. One such re-ranker is the exponential decay re-ranker. This re-ranker scales the model scores of items from the same creator shown in the same feed session by a power of a decay factor $\alpha < 1$. 

More specifically, the $k$-th item from the same creator (ranked by the second pass ranking score) will have its score scaled by $\alpha^{i-1}$, thus decreasing its ranking score. The online A/B test has several treatment groups with exponential decay factors ranging from 0.6 to 0.9. Larger decay factors generally bring user engagement gains. However, this does not mean the larger decay factors, the better user engagement. We conjecture that there is a creator diversity preference (in fact, personal feed composition preference beyond creator diversity) which requires multislot/whole page optimization.

\subsection{Offline Experiments}
As shown in Table \ref{tab:auc}, MultiSlot ReRanker models achieve significant AUC lift compared to models currently deployed in production. Our ablation study shows that interaction features among the current slot and previous slots play a key role. Without interaction features, the MultiSlot Logistic Regression (LR) model has smaller AUC lift, ranging from $+0.40\%$ to $+1.02\%$.
\begin{table}[]
    \centering
    \caption{Relative AUC Improvement By MultiSlot Models}
    \begin{tabular}{|c|c|c|} 
        \hline
        Response & MultiSlot Logistic Regression & MultiSlot XGBoost \\
        \hline like & $+3.86 \%$ & $+10.07 \%$ \\
        \hline skip & $+3.87 \%$ & $+6.82 \%$ \\
        \hline share & $+0.40 \%$ & $+8.08 \%$ \\
        \hline click & $+4.17 \%$ & $+8.50 \%$ \\
        \hline comment & $+0.41 \%$ & $+6.05 \%$ \\
        \hline
    \end{tabular}
    \label{tab:auc}
\end{table}

\begin{figure}
  \centering
  \includegraphics[scale=0.8]{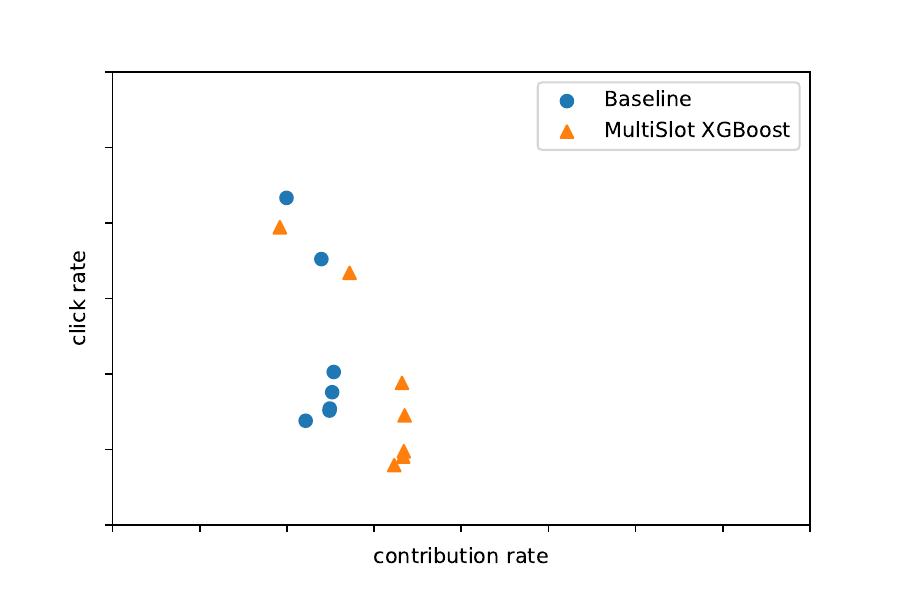}
  \caption{Replay Result of MultiSlot ReRanker XGBoost.}
  \label{fig:xgboost_replay}
\end{figure}

\begin{figure}
  \centering
  \includegraphics[scale=0.8]{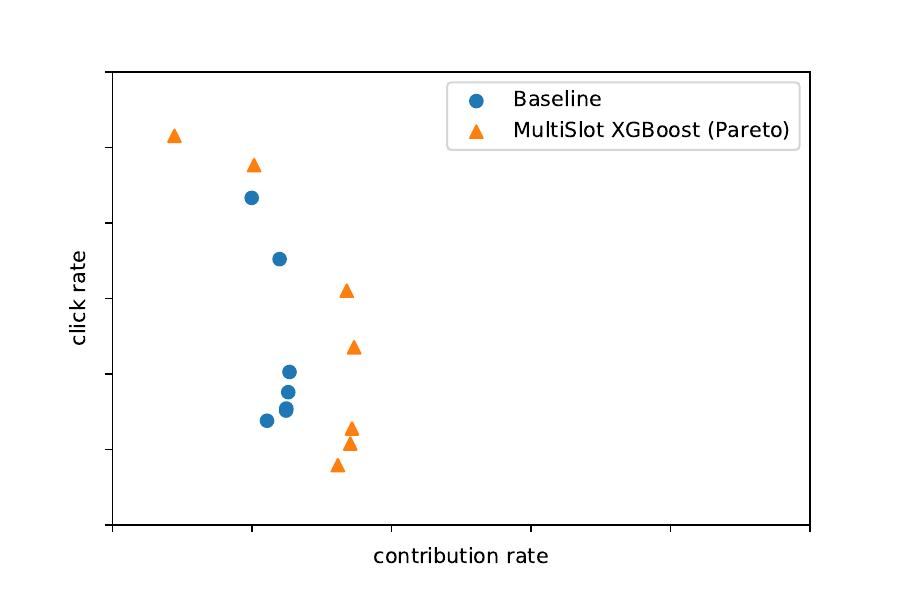}
  \caption{Replay Result of MultiSlot ReRanker XGBoost with Pareto Optimality.}
  \label{fig:xgboost_pareto_replay}
\end{figure}

The offline replay results of MultiSlot ReRanker models are shown in Figure \ref{fig:xgboost_replay} and Figure \ref{fig:xgboost_pareto_replay}. The exact number is omitted because of business confidentiality. Both curves demonstrate the trade-off among click and contribution responses (like/comment/share). MultiSlot ReRanker XGBoost model replay curve encompasses the baseline model, except in the high click rate region in Figure \ref{fig:xgboost_replay}. By keeping the second pass ranking click model and MultiSlot ReRanker contribution responses, the replay curve (Pareto Frontier) fully encompasses the baseline model in Figure \ref{fig:xgboost_pareto_replay}.

\subsection{Online Ramping}
We are working on the online A/B test. There are several challenges and learning lessons:
\begin{itemize}
    \item It matters where to put the MultiSlot ReRanker as there are multiple re-rankers existing in our online production system. For example, there are some rule-based re-rankers, ads slotting, as well as some revenue-related re-rankers, etc.
    \item As an extra re-ranker, MultiSlot ReRanker introduces extra latency. Because of strict latency requirement, we are limiting the total number of slots for multislot re-ranking process. 
    \item MultiSlot ReRanker should not completely overwrite the order imposed by the second pass ranking (SPR) model. The Sequential Greedy Algorithm \ref{alg:sequential_greedy} guarantees that the top slot is not changed, and each subsequent slot should not deviate from its original SPR order by more than 3 positions (the exact number can be tuned).
\end{itemize}

\section{MultiSlot Simulator}
The MultiSlot simulator is implemented as an OpenAI Gym Package\cite{brockmanOpenAIGym2016}. It can be integrated with RLLib/Ray\cite{liangRLlibAbstractionsDistributed2018}. It is a simulator environment for fast evaluation of potential algorithms. It can serve as a necessary condition for undertaking more complex approaches. The goal is not to learn perfect policies for production deployment. It simulates the re-ranking of an ordered list of documents from the previous scoring stage. The "sequential interaction" means that a user scans the list from top to bottom and interacts with items in sequence. 

It simulates the user choice model to generate the click label which is used to calculate the final reward\cite{ieRecSimConfigurableSimulation2019b}. One example of user choice model is Equation \ref{eqn:model_lr} where weights $w$ are pre-defined and fixed (i.e., "oracle" model). It also generates a list of items, along with SPR scores, item features such as item type and embedding. One method of generating the item features involves sampling from a given distribution such as uniform distributions within a chosen range appropriate. All the items in the list are sorted by the SPR score. 

An extensive comparison of both reinforcement learning (PPO\cite{schulmanProximalPolicyOptimization2017}, DQN\cite{mnihPlayingAtariDeep2013}, Policy Gradient\cite{suttonPolicyGradientMethods1999}) and supervised learning algorithms is shown in Table \ref{table:rl_sl_simulation}. It considers current slot's interactions with three previous slots. The top $K$ candidate at each re-ranking is $3$. It has been shown that:
\begin{itemize}
\item Most policies perform better than the $pointwise\_greedy$ policy, which approximates the behavior of the current model in production. 
\item The estimated greedy policy $sequential\_greedy\_estimated$ seems to be a good candidate for experiments on vertical products, as it performs better than other policies. 
\item By removing some features, the oracle model structure is partially known. In this case, the $sequential\_greedy\_estimated$ policy still performs better than the existing pointwise models.
\item Naive off-policy RL does better than pointwise greedy, but not as good as supervised models such as $sequential\_greedy\_estimated$. Part of the reason may be that the simulation environment may be too simple to have a good RL optimum solution. We see evidence of that when comparing oracle two-step receding horizon policies against oracle one-step greedy policies, which show similar performance.
\item RL models may perform better if there are changes as the parameters of the simulation are changed, which indicates that with another MultiSlot scenario, this may not hold. 
\end{itemize}

\begin{table}
\caption{An Overview of Reinforcement Learning and Supervised Learning Algorithms' Performance}
\label{table:rl_sl_simulation}
\setlength\tabcolsep{0pt}
\begin{tabular*}{\linewidth}{@{\extracolsep{\fill}} |c|c|l|c|c| }

\hline Policy & \begin{tabular}{l} 
On/Off policy
\end{tabular} & Policy Description & \begin{tabular}{l} 
Average \\
Reward
\end{tabular} & \begin{tabular}{l} 
Standard \\
Deviation
\end{tabular} \\
\hline random policy & N/A & \begin{tabular}{l} 
At each time step, we randomly select actions \\
$=1,2$ or 3
\end{tabular} & 2.3977 & 0.0112 \\
\hline \begin{tabular}{l} 
pointwise\_greedy \\
policy
\end{tabular} & N/A & \begin{tabular}{l} 
At each time step, we select the top item \\
based on the score from the previous layer. \\
This mimics the ranking models without a model \\
based re-ranker
\end{tabular} & 2.4326 & 0.0108 \\
\hline \begin{tabular}{c} 
sequential\_greedy \\
with oracle user choice model
\end{tabular} & N/A & \begin{tabular}{l} 
At each time step, we select the top scored \\
item based on the oracle user choice model. \\
Oracle user choice model uses the same \\
features + coefficients as used in the simulator
\end{tabular} & 2.7331 & 0.0119 \\
\hline \begin{tabular}{c} 
sequential\_greedy \\
with estimated user choice model
\end{tabular} & N/A & \begin{tabular}{l} 
We first generate data from the oracle choice \\
model, and train a model to estimate its \\
parameters. Then, at each time step, we \\
select the top scored item based on the \\
estimated model.
\end{tabular} & 2.7089 & 0.0119 \\
\hline DQN & \begin{tabular}{l} 
Off-policy from \\
random data
\end{tabular} & \begin{tabular}{l} 
Best policy from simple hyperparamter sweep \\
with the same features as user choice model
\end{tabular} & 2.587 & 0.0212 \\
\hline PPO & \begin{tabular}{l} 
Off-policy from \\
random data \\
(PPO is \\
typically on \\
policy)
\end{tabular} & \begin{tabular}{l} 
Best policy from simple hyperparamter sweep \\
with the same features as user choice model
\end{tabular} & 2.6307 & 0.0222 \\
\hline Policy Gradient & On-policy data & \begin{tabular}{l} 
Best policy from simple hyperparamter sweep \\
with the same features as user choice model
\end{tabular} & 2.61433 & 0.0222 \\
\hline
\end{tabular*}
\end{table}

The MultiSlot simulator is also used to validate the one-step importance sampling offline replay. It samples both actual rewards (ground truth), one-step rewards, pointwise greedy, and sequential greedy, etc. The one-step method mostly captures the relative ordering which provides some confidence in its capability.

\section{Conclusions and Future Work}
In this paper, we propose a new generic model-based re-ranking framework in recommendation systems. We also propose a new method for multislot offline replay which is validated in both simulator and production dataset. The MultiSlot simulator based on OpenAI Gym can be easily configured for different assumptions and scenarios. It creates opportunities for exploring new solutions to the multislot whole page re-ranking optimization problem.

For future work, we are interested in using multi-step optimization to solve the MultiSlot ReRanker problem. The goal is to determine the optimal action which maximizes the total reward. The reward can be defined as a linear combination of contributions (like/comment/share) and click. 

Another direction is to use action value function methods which are lightweight extensions of supervised models to predict multi-step response. Action value functions model the expected reward over the next few steps as a function of the current action taken. For the next few steps, it is assumed that we follow the existing policy. Simply put, the training data for such a policy at slot $i$ is
 $\left(s_{i}, r_i+\lambda r_{i+1}+\lambda r_{i+2}+\ldots\right)$ where $(0<=\lambda<=1)$ is a decay factor. This model measures the impact of an action on multiple future steps. One step greedy policies, which choose the action that maximizes this reward at each step, are probably better than data collection policies, according to the theory derived in policy iteration (particularly, it forms the policy improvement step). Note that the action value function is a supervised model, similar to the Sequential Greedy Algorithm \ref{alg:sequential_greedy}. It just uses a modified label.

\textbf{Acknowledgement.} We would like to thank Ying Xuan, Samaneh Moghaddam and LinkedIn Feed AI team for the collaboration, thank Feed Infra team and Borja Ocejo Elizondo for their efforts and supports in online A/B testing, Gaurav Srivastava and Keerthi Selvaraj for the optimizer experiments, Sarah Xing for discussions in the early stage, Aman Gupta for discussions about online results. We would like to thank Lu Chen, David Golland, Lu Zheng, Katherine Vaiente and Jon Adams for reviewing this paper.

\printbibliography

@online{belloSeq2SlateRerankingSlate2019a,
  title = {{{Seq2Slate}}: {{Re-ranking}} and {{Slate Optimization}} with {{RNNs}}},
  shorttitle = {{{Seq2Slate}}},
  author = {Bello, Irwan and Kulkarni, Sayali and Jain, Sagar and Boutilier, Craig and Chi, Ed and Eban, Elad and Luo, Xiyang and Mackey, Alan and Meshi, Ofer},
  date = {2019-03-19},
  eprint = {1810.02019},
  eprinttype = {arxiv},
  eprintclass = {cs, stat},
  doi = {10.48550/arXiv.1810.02019},
  url = {http://arxiv.org/abs/1810.02019},
  urldate = {2023-10-30},
  pubstate = {preprint}
}

@online{brockmanOpenAIGym2016,
  title = {{{OpenAI Gym}}},
  author = {Brockman, Greg and Cheung, Vicki and Pettersson, Ludwig and Schneider, Jonas and Schulman, John and Tang, Jie and Zaremba, Wojciech},
  date = {2016-06-05},
  eprint = {1606.01540},
  eprinttype = {arxiv},
  eprintclass = {cs},
  doi = {10.48550/arXiv.1606.01540},
  url = {http://arxiv.org/abs/1606.01540},
  urldate = {2023-10-30},
  pubstate = {preprint}
}

@inproceedings{chenTopKOffPolicyCorrection2019,
  title = {Top-{{K Off-Policy Correction}} for a {{REINFORCE Recommender System}}},
  author = {Chen, Minmin and Beutel, Alex and Covington, Paul and Jain, Sagar and Belletti, Francois and Chi, Ed},
  date = {2019},
  url = {https://arxiv.org/pdf/1812.02353.pdf},
  urldate = {2023-11-02}
}

@inproceedings{chenXGBoostScalableTree2016,
  title = {{{XGBoost}}: {{A Scalable Tree Boosting System}}},
  shorttitle = {{{XGBoost}}},
  booktitle = {Proceedings of the 22nd {{ACM SIGKDD International Conference}} on {{Knowledge Discovery}} and {{Data Mining}}},
  author = {Chen, Tianqi and Guestrin, Carlos},
  date = {2016-08-13},
  eprint = {1603.02754},
  eprinttype = {arxiv},
  eprintclass = {cs},
  pages = {785--794},
  doi = {10.1145/2939672.2939785},
  url = {http://arxiv.org/abs/1603.02754},
  urldate = {2023-11-02}
}

@inproceedings{covingtonDeepNeuralNetworks2016,
  title = {Deep {{Neural Networks}} for {{YouTube Recommendations}}},
  booktitle = {Proceedings of the 10th {{ACM Conference}} on {{Recommender Systems}}},
  author = {Covington, Paul and Adams, Jay and Sargin, Emre},
  date = {2016},
  location = {{New York, NY, USA}}
}

@inproceedings{dingWholePageOptimization2019a,
  title = {Whole {{Page Optimization}} with {{Global Constraints}}},
  booktitle = {Proceedings of the 25th {{ACM SIGKDD International Conference}} on {{Knowledge Discovery}} \& {{Data Mining}}},
  author = {Ding, Weicong and Govindaraj, Dinesh and Vishwanathan, S V N},
  date = {2019-07-25},
  series = {{{KDD}} '19},
  pages = {3153--3161},
  publisher = {{Association for Computing Machinery}},
  location = {{New York, NY, USA}},
  doi = {10.1145/3292500.3330675},
  url = {https://doi.org/10.1145/3292500.3330675},
  urldate = {2023-10-30},
  isbn = {978-1-4503-6201-6}
}

@online{ieRecSimConfigurableSimulation2019b,
  title = {{{RecSim}}: {{A Configurable Simulation Platform}} for {{Recommender Systems}}},
  shorttitle = {{{RecSim}}},
  author = {Ie, Eugene and Hsu, Chih-wei and Mladenov, Martin and Jain, Vihan and Narvekar, Sanmit and Wang, Jing and Wu, Rui and Boutilier, Craig},
  date = {2019-09-26},
  eprint = {1909.04847},
  eprinttype = {arxiv},
  eprintclass = {cs, stat},
  doi = {10.48550/arXiv.1909.04847},
  url = {http://arxiv.org/abs/1909.04847},
  urldate = {2023-10-30},
  pubstate = {preprint}
}

@inproceedings{ieSlateQTractableDecomposition2019a,
  title = {{{SlateQ}}: {{A Tractable Decomposition}} for {{Reinforcement Learning}} with {{Recommendation Sets}}},
  shorttitle = {{{SlateQ}}},
  booktitle = {Proceedings of the {{Twenty-eighth International Joint Conference}} on {{Artificial Intelligence}} ({{IJCAI-19}})},
  author = {Ie, Eugene and Jain, Vihan and Wang, Jing and Narvekar, Sanmit and Agarwal, Ritesh and Wu, Rui and Cheng, Heng-Tze and Chandra, Tushar and Boutilier, Craig},
  date = {2019},
  pages = {2592--2599},
  location = {{Macau, China}}
}

@online{jiangGreedyRankingSlate2019,
  title = {Beyond {{Greedy Ranking}}: {{Slate Optimization}} via {{List-CVAE}}},
  shorttitle = {Beyond {{Greedy Ranking}}},
  author = {Jiang, Ray and Gowal, Sven and Mann, Timothy A. and Rezende, Danilo J.},
  date = {2019-02-23},
  eprint = {1803.01682},
  eprinttype = {arxiv},
  eprintclass = {cs, stat},
  doi = {10.48550/arXiv.1803.01682},
  url = {http://arxiv.org/abs/1803.01682},
  urldate = {2023-10-30},
  pubstate = {preprint}
}

@online{liangRLlibAbstractionsDistributed2018,
  title = {{{RLlib}}: {{Abstractions}} for {{Distributed Reinforcement Learning}}},
  shorttitle = {{{RLlib}}},
  author = {Liang, Eric and Liaw, Richard and Moritz, Philipp and Nishihara, Robert and Fox, Roy and Goldberg, Ken and Gonzalez, Joseph E. and Jordan, Michael I. and Stoica, Ion},
  date = {2018-06-28},
  eprint = {1712.09381},
  eprinttype = {arxiv},
  eprintclass = {cs},
  doi = {10.48550/arXiv.1712.09381},
  url = {http://arxiv.org/abs/1712.09381},
  urldate = {2023-10-30},
  pubstate = {preprint}
}

@inproceedings{liUnbiasedOfflineEvaluation2011,
  title = {Unbiased {{Offline Evaluation}} of {{Contextual-bandit-based News Article Recommendation Algorithms}}},
  booktitle = {Proceedings of the Fourth {{ACM}} International Conference on {{Web}} Search and Data Mining},
  author = {Li, Lihong and Chu, Wei and Langford, John and Wang, Xuanhui},
  date = {2011-02-09},
  eprint = {1003.5956},
  eprinttype = {arxiv},
  eprintclass = {cs, stat},
  pages = {297--306},
  doi = {10.1145/1935826.1935878},
  url = {http://arxiv.org/abs/1003.5956},
  urldate = {2023-11-02}
}

@online{mnihPlayingAtariDeep2013,
  title = {Playing {{Atari}} with {{Deep Reinforcement Learning}}},
  author = {Mnih, Volodymyr and Kavukcuoglu, Koray and Silver, David and Graves, Alex and Antonoglou, Ioannis and Wierstra, Daan and Riedmiller, Martin},
  date = {2013-12-19},
  eprint = {1312.5602},
  eprinttype = {arxiv},
  eprintclass = {cs},
  doi = {10.48550/arXiv.1312.5602},
  url = {http://arxiv.org/abs/1312.5602},
  urldate = {2023-11-02},
  pubstate = {preprint}
}

@online{peiPersonalizedRerankingRecommendation2019a,
  title = {Personalized {{Re-ranking}} for {{Recommendation}}},
  author = {Pei, Changhua and Zhang, Yi and Zhang, Yongfeng and Sun, Fei and Lin, Xiao and Sun, Hanxiao and Wu, Jian and Jiang, Peng and Ou, Wenwu},
  date = {2019-07-22},
  eprint = {1904.06813},
  eprinttype = {arxiv},
  eprintclass = {cs},
  doi = {10.48550/arXiv.1904.06813},
  url = {http://arxiv.org/abs/1904.06813},
  urldate = {2023-10-30},
  pubstate = {preprint}
}

@online{schulmanProximalPolicyOptimization2017,
  title = {Proximal {{Policy Optimization Algorithms}}},
  author = {Schulman, John and Wolski, Filip and Dhariwal, Prafulla and Radford, Alec and Klimov, Oleg},
  date = {2017-08-28},
  eprint = {1707.06347},
  eprinttype = {arxiv},
  eprintclass = {cs},
  doi = {10.48550/arXiv.1707.06347},
  url = {http://arxiv.org/abs/1707.06347},
  urldate = {2023-11-02},
  pubstate = {preprint}
}

@inproceedings{suttonPolicyGradientMethods1999,
  title = {Policy {{Gradient Methods}} for {{Reinforcement Learning}} with {{Function Approximation}}},
  booktitle = {Advances in {{Neural Information Processing Systems}}},
  author = {Sutton, Richard S and McAllester, David and Singh, Satinder and Mansour, Yishay},
  date = {1999},
  volume = {12},
  publisher = {{MIT Press}},
  url = {https://papers.nips.cc/paper_files/paper/1999/hash/464d828b85b0bed98e80ade0a5c43b0f-Abstract.html},
  urldate = {2023-11-02}
}

@inproceedings{zhaoDeepReinforcementLearning2018a,
  title = {Deep {{Reinforcement Learning}} for {{Page-wise Recommendations}}},
  booktitle = {Proceedings of the 12th {{ACM Conference}} on {{Recommender Systems}}},
  author = {Zhao, Xiangyu and Xia, Long and Zhang, Liang and Ding, Zhuoye and Yin, Dawei and Tang, Jiliang},
  date = {2018-09-27},
  eprint = {1805.02343},
  eprinttype = {arxiv},
  eprintclass = {cs},
  pages = {95--103},
  doi = {10.1145/3240323.3240374},
  url = {http://arxiv.org/abs/1805.02343},
  urldate = {2023-10-30}
}

\end{document}